\documentclass[sigconf]{acmart}

\AtBeginDocument{%
  \providecommand\BibTeX{{%
    \normalfont B\kern-0.5em{\scshape i\kern-0.25em b}\kern-0.8em\TeX}}}
    
\usepackage{helvet} 
\usepackage{courier}  
\usepackage{graphicx} 
\usepackage{url}
\usepackage{multirow}
\usepackage{bm}
\usepackage{algorithm}
\usepackage{algpseudocode}  
\usepackage{longtable}
\usepackage{array}
\usepackage{caption}
\usepackage{subfigure}
\usepackage{natbib}
\usepackage{balance}
\usepackage{microtype}

\copyrightyear{2020} 
\acmYear{2020} 
\setcopyright{acmcopyright}
\acmConference[MM '20]{Proceedings of the 28th ACM International Conference on Multimedia}{October 12--16, 2020}{Seattle, WA, USA}
\acmBooktitle{Proceedings of the 28th ACM International Conference on Multimedia (MM '20), October 12--16, 2020, Seattle, WA, USA}
\acmPrice{15.00}
\acmDOI{10.1145/3394171.3413747}
\acmISBN{978-1-4503-7988-5/20/10}

\begin{document}
\fancyhead{}



\title{Cognitive Representation Learning of Self-Media Online Article Quality}

\author{Yiru Wang}
\affiliation{%
  \institution{Data Quality Team, WeChat, Tencent Inc. \& Tsinghua Shenzhen International Graduate School}
  \streetaddress{No.33,Haitian 2 Road}
  \city{Shenzhen}
  \state{Guangdong}
  \country{PR China}
  \postcode{518054}
}
\email{wangyiru017@gmail.com}

\author{Shen Huang}
\affiliation{%
  \institution{WeChat, Tencent Inc.}
  \streetaddress{No.33,Haitian 2 Road}
  \city{Shenzhen}
  \state{Guangdong}
  \country{PR China}}
\email{shinhuang@tencent.com}

\author{Gongfu Li}
\affiliation{%
  \institution{WeChat, Tencent Inc.}
  \city{Shenzhen}
  \state{Guangdong}
  \country{PR China}
}
\email{gongfuli@tencent.com}

\author{Qiang Deng}
\affiliation{%
 \institution{WeChat, Tencent Inc.}
  \city{Shenzhen}
  \state{Guangdong}
  \country{PR China}}
\email{calvindeng@tencent.com}

\author{Dongliang Liao}
\affiliation{%
  \institution{WeChat, Tencent Inc.}
  \city{Shenzhen}
  \state{Guangdong}
  \country{PR China}}
 \email{brightliao@tencent.com}

\author{Pengda Si}
\affiliation{%
  \institution{Tsinghua Shenzhen International Graduate School, Tsinghua University}
  \city{Shenzhen}
  \state{Guangdong}
  \country{PR China}
  \postcode{518055}}
\email{spd18@mails.tsinghua.edu.cn}

\author{Yujiu Yang}
\authornote{Both authors are corresponding authors.}
\affiliation{\institution{Tsinghua Shenzhen International Graduate School, Tsinghua University}
  \city{Shenzhen}
  \state{Guangdong}
  \country{PR China}}
\email{yang.yujiu@sz.tsinghua.edu.cn}
\author{Jin Xu}
\affiliation{\institution{Data Quality Team, WeChat, Tencent Inc.}
  \city{Shenzhen}
  \state{Guangdong}
  \country{PR China}}
\authornotemark[1]
\email{jinxxu@tencent.com}



\begin{abstract}
  The automatic quality assessment of self-media online articles is an urgent and new issue, which is of great value to the online recommendation and search. Different from traditional and well-formed articles, self-media online articles are mainly created by users, which have the appearance characteristics of different text levels and multi-modal hybrid editing, along with the potential characteristics of diverse content, different styles, large semantic spans and good interactive experience requirements. To solve these challenges, we establish a joint model CoQAN in combination with the layout organization, writing characteristics and text semantics, designing different representation learning subnetworks, especially for the feature learning process and interactive reading habits on mobile terminals. It is more consistent with the cognitive style of expressing an expert's evaluation of articles. We have also constructed a large scale real-world assessment dataset. Extensive experimental results show that the proposed framework significantly outperforms state-of-the-art methods, and effectively learns and integrates different factors of the online article quality assessment.
\end{abstract}




\maketitle

\section{Introduction}
In the era of mobile reading, a lot of self-media platforms based on the user-generated-content mode have emerged. People are accustomed to spending fragmented time reading online articles published on self-media platforms to conveniently get information and knowledge via mobile devices. Different from traditional documents such as essays, academic papers or Wikipedia documents, self-media online articles have more diverse multimedia elements, in addition to text, usually existing pictures, videos, etc. The organization of these elements jointly affects users' perception. Besides, since the creation forms of self-media online articles are more free, these articles do not have a unified format and layout, and usually vary in diverse categories, styles and content. Therefore, it is necessary to integrate different multimedia elements more comprehensively to jointly process self-media online articles.

The openness of self-media platforms, where each user can be a producer, however, results in uneven quality of online articles. Assessing the quality of self-media online articles is a critical issue for many applications such as recommender systems and online search to find high-quality articles and filter low-quality articles. It is very helpful to increase user stickiness to propose an efficient solution for the automatic evaluation of the self-media online article quality. Considering the nature of self-media platforms, in order to engage users, the quality of self-media online articles is reasonably defined as the level of the reading experience that articles give users. This can be reflected in the article's content, writing norms, user perception, etc., and each factor also contains complicated elements, making the self-media online article quality assessment a much more complex and challenging task. The following question lies to be addressed: \textbf{How to establish a unified framework to effectively solve the multivariate representation learning of self-media online article quality?} However, current studies on the document quality assessment mainly focus on textual features \citep{TaghipourN16,DongZY17}. To the best of our knowledge, no prior work has studied the automatic self-media online article quality assessment. 


As we know, the cognitive process while human read and qualitatively evaluate a self-media online article is from the surface to the centre. When a reader clicks an article, the first thing the reader feels is the layout appearance, which is the reader's surface cognition. Nice visual layouts and rich presentational forms can make readers interested in the article and give readers a better reading experience. Then the reader gets the main impression of the content by browsing the vocabulary, syntax, article organization and pictures, which is the reader's shallow cognition. Finally, the reader needs a deep understanding of semantics and logic to appreciate the sense and value of the article, which is the reader's deep cognition. These three cognitive levels respectively correspond to the following important qualitative properties of online articles form the surface to the deep cognition: 
\begin{itemize}
\item \textbf{Layout Organization:} It interprets how an article is arranged by content blocks. Each block consists of consecutive homogeneous content, denoting a single idea or information, such as a textual paragraph or an image. The layout organization not only considers the presentational patterns of the image and text arrangement, but also reflects the level of the content organization.

\item \textbf{Writing Characteristics:} It interactively fuses multi-modal features to get an overall content impression of an article, considering the image quality and perception such as the image clarity and textual area proportion in the image, also overviewing the vocabulary, syntax and grammatical elements of the content and title, along with organization styles such as the number of sections, subtitles and pictures.

\item \textbf{Text Semantics:} It refers to the text content understanding and writing logic analysis, which needs to deeply learn the document-level coherence and co-reference among words and sentences.
\end{itemize}

Inspired by the cognitive process above, we propose to assess the self-media online article quality in combination with the layout organization, writing characteristics and text semantics, interacting feature learning with each other and integrated into a unified model framework. Most self-media online articles mainly convey the core ideas through text. For our task, the quality and visual perception of pictures are more important than picture semantics. Thus, we use pictures as the key units to model the page layouts, and important picture features are extracted to reflect the visual perception level and the readability of articles in our proposed method. For layout modeling, following the top-to-bottom sequential reading habits of people, we design a layout organization subnetwork to explicitly learn the arrangement patterns of the content block sequence. Besides, we design a writing characteristics subnetwork which aims to perform deep feature selection and feature fusion, rather than a combination of first-order or low-order features in traditional models. In addition, we also design a text semantics subnetwork to deeply capture the text semantics and cohesive relationships within the entire article from the perspective of different semantic levels. Finally, these three subnetworks together constitute our joint model \textbf{CoQAN}. Our CoQAN considers multiple important factors affecting online article quality to systematically and comprehensively assess the quality from a multivariate perspective. Unfortunately, although there are more and more self-media user generated articles, no such dataset is available for research. So, we have constructed a real-world dataset for the self-media online article quality assessment. In general, our contributions are summarized as follows:
\begin{itemize}
\item To the best of our knowledge, this is the first study to solve the automatic quality assessment of self-media online articles. The proposed approach can well model the scoring elements and reading habits of human experts.
\item We further propose a joint model combining high-order feature representation learning and different characterization subnetworks, and construct an end-to-end framework for the self-media online article quality assessment.
\item We construct a large scale real-world dataset. Extensive experimental results show that the proposed model significantly outperforms state-of-the-art methods.
\end{itemize}

\begin{figure*}[ht]
\centerline{\includegraphics[width=14cm,height=7.8cm]{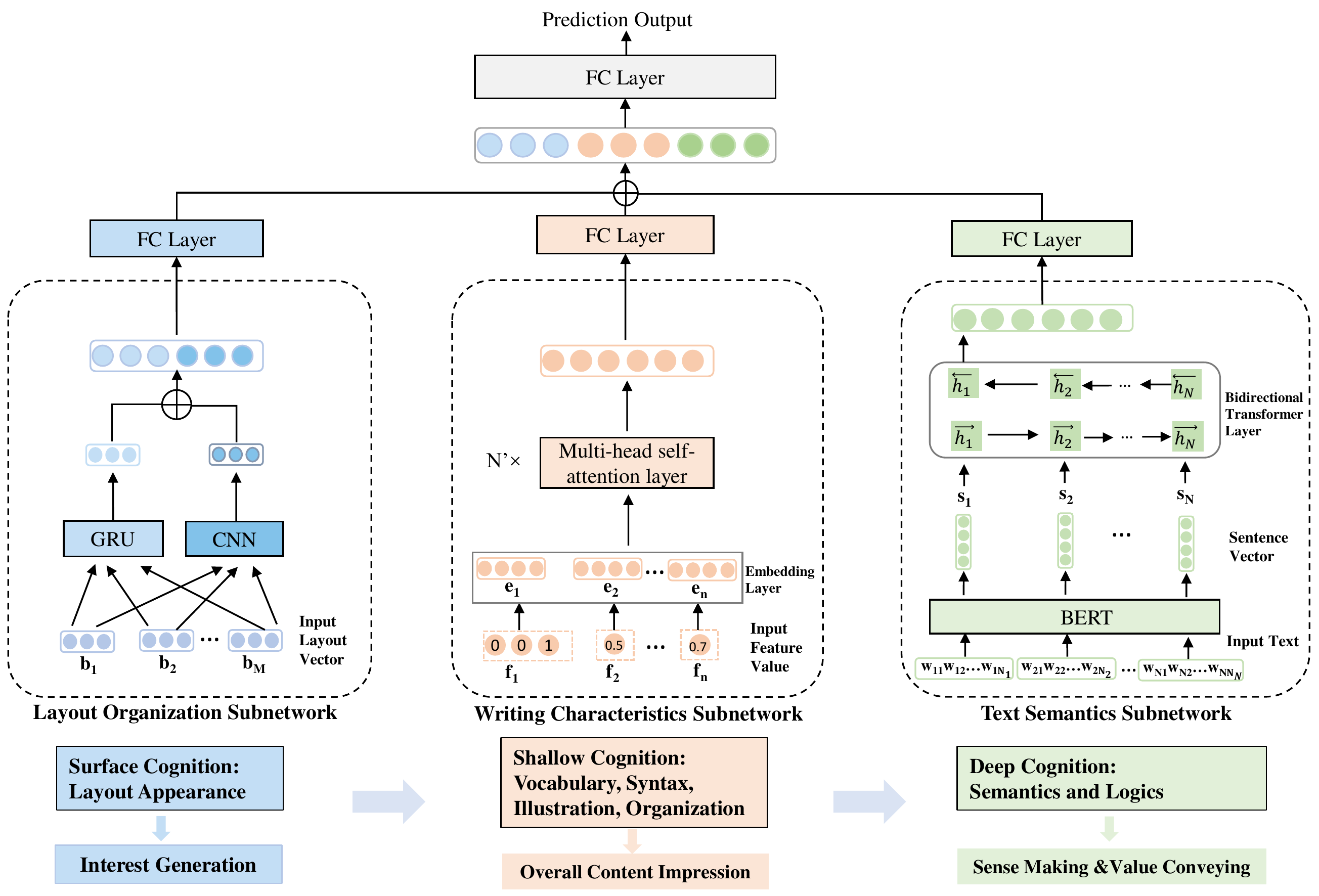}}
\caption{The architecture of our cognitive joint network CoQAN for the self-media online article quality classification.}
\label{fig:framework}
\end{figure*}
\section{Related Work}
Some works are conducted for the document quality assessment across relevant domains: Automatic Essay Scoring (AES), Wikipedia Document Quality Assessment and Academic Paper Rating.

\textbf{Handcrafted feature based methods.}~ Besides traditional features such as n-grams, part-of-speech tags, word error rate and essay length, the lexical and semantic overlap \citep{PersingN14, PhandiCN15}, syntactic and sentence structure features \citep{CumminsZB16, ChenH13}, text coherence and sentence cohesion \citep{ChenH13} have been designed and explored in AES. For Wikipedia documents, some meta features such as the number of headings, images and references \citep{Warncke-WangAHT15}, the number and authority of editors \citep{WangI11, SteinH07} are leveraged. \citet{KangADZKHS18} predicted whether an academic paper is accepted or rejected based on features such as the title length and whether specific words appear in the abstract. Besides the drawbacks of time consuming feature engineering and data sparsity, these feature-based approaches rarely consider the multi-modal features, without fully exploiting the rich stylistic and semantic information, and unable to model deep feature fusion spaces to learn more complex contextual dependencies.

\textbf{Neural network based methods.}  Most existing deep learning based works use recurrent neural networks (RNN) and convolutional neural networks (CNN) to model the input text, generating a single representation vector of the text for prediction \citep{AlikaniotisYR16,TaghipourN16}. \citet{Fei2016} developed a hierarchical CNN model for AES task by using two CNN on both sentence level and text level. Similarly, \citet{DongZY17} utilized a CNN to obtain the sentence representation and an LSTM to obtain the essay representation, with attention pooling at these two levels. \citet{YangSLM18} exploited a modularized hierarchical CNN for the academic paper quality classification. Although neural network based methods perform better than traditional statistical methods, they fail to consider integral linguistic and cognitive factors in documents, which play an important role in document quality assigned by experts \citep{DasguptaNDS18}. 

A few studies improve deep learning models by combining with linguistic and cognitive features. \citet{DasguptaNDS18} augmented different linguistic, cognitive and psychological features of text along with a hierarchical convolution recurrent neural network framework for AES. \citet{BhattacharyyaKP18} used cognitive information obtained from reader's gaze behavior to help predict the score the user would assign to the text quality. \citet{ShenSB019} combined textual features with a visual rendering of the document by fine-tuning an Inception V3 model for the Wikipedia document and academic paper quality classification. However, the way to process visual renderings can't explicitly extract and capture the features we expect, meanwhile the deep feature combination of convolutions is uncontrollable. Moreover, according to our reading habits, layout information is viewed from top to bottom, the translational invariance of the convolution operation fails to model this important sequential information.


\section{Methodology}
\subsection{Task Definition and Model Overview}
We treat the self-media online article quality assessment as a classification task, i.e., given an article, we predict whether it should be regarded as a high-quality article or a low-quality one. The article quality in this paper is defined as the level of the reading experience that articles bring to users, which largely depends on the readability of articles, involving appearance characteristics, writing norms and content semantics. For high-quality articles, the layout is neat and beautiful, with clear paragraphs and sections; pictures and text are well-arranged; the content is coherent and cohesive, with good writing logic and rich information. Conversely, for low-quality articles, the layout or writing logic is confusing, or the content is incomplete or meaningless, even maybe a piece of text crowded together or messy pure images. 

After partitioning among paragraphs and images or videos, we can obtain a layout structure sequence $B=\{b_1,b_2,\ldots, b_M\}$ of an online article, where each $b_i$ represents a content block which may be a subtitle, paragraph, picture or video. For the text in an online article, it is as a sequence of $N$ sentences $S=\{s_1,s_2,\ldots,s_N\}$, and each $s_i$ as a sequence of $N_i$ tokens $s_i=\{w_{i1},w_{i2},\ldots,w_{iN_i}\}$, where $w_{ij}$ represents the token at position $j$ in sentence $i$. $F=\{f_1,f_2,\ldots,f_n\}$ denotes a set of feature fields of an article, containing $n$ different features related to the writing and organization styles. Our task is to predict the true quality class label $y$ on the condition of $D=\{B,F,S\}$ for an online article.

Figure \ref{fig:framework} demonstrates the architecture of our proposed joint model CoQAN. We design three subnetworks to decouple the modeling of the layout organization, writing characteristics and text semantics. The three subnetworks are aggregated into a unified model in a flexible way that each subnetwork passes through a fully connected layer to adjust the weight of each neuron in its output vector. After this, three intermediate vectors are obtained and cascaded together to output a predicted classification label. 

\subsection{Layout Organization Subnetwork}
\begin{figure}[ht]
\centerline{\includegraphics[width=\columnwidth,height=4cm]{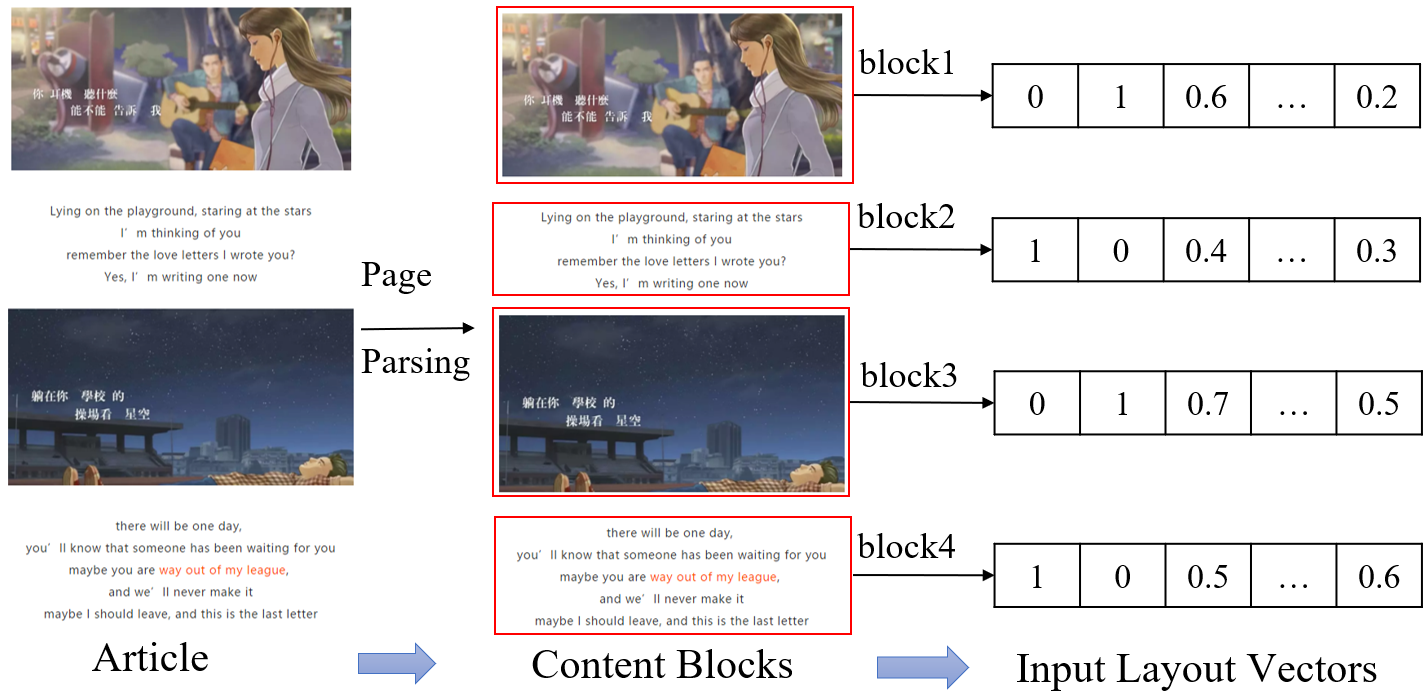}}
\caption{Content block extraction and input feature vector construction in the layout organization subnetwork.}
\label{fig:layout}
\end{figure}

In order to explicitly learn the organization and arrangement of presentational layouts, we first partition an article into a sequence of content blocks by page parsing. Each content block can be a subtitle, paragraph, picture or video, as shown in Figure \ref{fig:layout}. After that, we extract the features related to the page layouts for each content block, such as the type (text/ picture/ video), position, height, distance from the top of the page, etc. Then we construct a feature vector $b_i$ for each content block, where $b_i\in R^m$ denotes the concatenation of these layout features, $m$ is the dimension of concatenated features. This allows our constructed feature vectors to express multiple important layout features, and each dimension is meaningful.

We employ a gated recurrent unit (GRU) \citep{ChoMGBBSB14} network on the input sequence $B=\{b_1,b_2,\ldots, b_M\}$ to model the sequence dependency among $M$ content blocks and capture the global arrangement pattern of the article. The GRU is a widely used RNN structure for its superior ability to capture long distance dependencies. The last hidden state $h_M$ is put as the output vector of this GRU network.


It is also important to capture local layout patterns. Thus we exploit 1-D convolutional neural network on the input layout vectors, which has been proved optimum for capturing local features. Moreover, we adopt multiple kernels with different sizes to capture different scales of layout patterns. A convolution operation involves a filter $k\in R^{hm}$ which is applied to a window of $h$ blocks to produce a new feature. Specifically, a feature $c_i^k$ is generated from a window of blocks $b_{i:i+h-1}$, and this filter is applied to each possible window of blocks in the sequence ${b_{1:h}, b_{2:h+1},\ldots ,b_{M-h+1:M}}$ to produce a feature map. We then apply a max pooling operation over the feature map as the feature $\tilde{c^k}$ corresponding to this particular filter $k$. We use $K$ filters with variable window sizes to obtain $K$ features. We concatenate these $K$ features as the output vector $h^c$ of this CNN network. 
\begin{gather}
\begin{split}
    h^c &= \tilde{c^1}\oplus \tilde{c^2} \oplus \ldots \oplus \tilde{c^K} \\
   \tilde{c^k} &= max\{c_1^k, c_2^k, \ldots, c_{M-h+1}^k\} \\
   c_i^k &= ReLU(w \cdot b_{i:i+h-1} + b ) 
\end{split}
\end{gather}
where $b$ is a bias term, $h$ is a variable window size. We concatenate the output vectors of the GRU and the CNN network to be the final vector $h^l= h_M \oplus h^c$ of our layout organization subnetwork.

\subsection{Writing Characteristics Subnetwork}
The writing characteristics subnetwork captures the editing style of an online article. We extract and calculate features representing the attributes of the online article as inputs, including features of the title, body, image, video and organizational features, etc. For example, title features include title length, keyword number, etc.; body features consist of article category, text length, n-grams, part-of-speech tags, unique character and word proportions, etc.; image and video features contain image number, GIF number, video number, maximum number of OCR characters, etc., and organizational features include paragraph number, the number of template images as the signal of sections, the ratio of images to paragraphs, etc.



Now, we have a set of features $F=\{f_1,f_2,\ldots,f_n\}$, where $f_i$ is a one-hot vector for a categorical field and a scalar value for a numerical field. To allow the interactions between categorical and numerical features, we project them into the same feature space by the embedding layer. Then we aim to learn high-order combinatorial features in the embedding space. The key problem is to determine which features should be selected and combined to form meaningful high-order features, that is traditionally accomplished by domain experts based on their knowledge or specific criteria \citep{XuTHM17}. In this paper, we tackle this problem by applying the multi-head self-attention layers \citep{abs} to model the correlations between different feature fields. We first define the correlation between arbitrary two feature embeddings $e(f_p)$ and $e(f_q)$ under a specific attention head $h$ as follows:
\begin{gather}
\begin{split}
   a_{pq}^h&=\frac{exp(\Psi_{pq}^h)}{\sum_{k=1}^nexp(\Psi_{pk}^h)} \\
   \Psi_{pq}^h&=< w_p^he(f_p),w_q^he(f_q)>
\end{split}
\end{gather}
where $\Psi_{pq}^h$ is a function which defines the similarity between feature embeddings $e(f_p)$ and $e(f_q)$, here we use the inner product to calculate the similarity due to its effectiveness. $w_p^h$ and $w_q^h$ are projection vectors for $e(f_p)$ and $e(f_q)$ respectively, and we use different $w_p^h$ and $w_q^h$ for different heads to project the original embedding space into multiple new subspaces. $a_{pq}^h$ is the attention weight value of feature $q$ to feature $p$ in the subspace $h$. Next, we produce the representation $r_i^h$ for each feature $i$ in the subspace $h$ via combining all relevant features guided by attention coefficients:
\begin{equation}
    r_i^h =\sum\nolimits_{j=1}^na_{ij}^h\cdot w_j^he(f_j) 
\end{equation}
where $w_j^h$ is the projection vector for the embedding $e(f_j)$. $r_i^h$ represents a new combinatorial feature fusing feature $i$ and its relevant features. Furthermore, we use multiple heads to allow a feature likely involved in different combinatorial features, which create different subspaces and learn distinct feature interactions separately. We collect combinatorial features learned in all subspaces to get the representation of high-order features involving feature $i$ as follow:
\begin{equation}
    r_i = ReLU((r_i^1\oplus r_i^2 \oplus \ldots \oplus r_i^H)+ w_ie(f_i))
\end{equation}
where $\oplus$ is the concatenation operator, $H$ is the total number of attention heads, $w_i$ is the projection vector, $e(f_i)$ is the embedding vector of $f_i$, and $ReLU$ is a non-linear activation function. We add standard residual connections in our network to preserve previously learned combinatorial features, including raw individual features.

We stack $N^{'}$ multiple above layers with the output of the previous multi-head self-attention layer as the input of the next one, that can model different orders of combinatorial features. After stacking layers, we get a set of feature vectors ${r_1^{'},r_2^{'},\ldots,r_n^{'}}$ corresponding to $n$ feature fields. Finally, we simply concatenate all of them to generate the final output vector $h^w =r_1^{'}\oplus r_2^{'} \oplus \ldots \oplus r_n^{'}$ of the writing characteristics subnetwork. 
\subsection{Text Semantics Subnetwork}
Since the text quality depends on deep semantics, the theme and content depth, designing an effective text learning network is especially important. Relying on the powerful pre-training technique and the ability to model contextual relationships, we use BERT \citep{bert} to encode raw text. But the problem is that the time complexity of the self-attention mechanism in BERT is $O(n^2)$ for the sequence with length $n$, so BERT is only used in the sentence-level modeling at present, not suitable for document-level long text inputs. In this regard, we improve BERT into a hierarchical structure called hi-BERT, considering the inherent hierarchical structure of documents, i.e. words form a sentence and sentences form a document. Hi-BERT encodes a document with two encoders applied at the sentence level and document level, respectively. 

At the sentence level, we use the pre-tained BERT model in a migration learning way, that we fine-tune the parameters oriented to our task during training. Since the title is the high-level overview of an article and shows the primary impression, the title is also as an input sentence. For each sentence $s_i$, the sentence-level encoder learns the sentence encoding vector $s_i^r$ on its word sequence.

At the document level, we hope to learn the deep interrelationships between sentences, so we employ a bidirectional Transformer \citep{transformer} layer which follows the block structure of BERT, taking the sequence of sentence encoding vectors as input. The header output vector of the bidirectional Transformer is finally set as the document vector $h^d$. Our hi-BERT can deeply learn interactive relationships between words and sentences, and generate a more robust document representation when facing different writing styles.
\begin{gather}
\begin{split}
    h^d &= \overrightarrow{h_{1}} \oplus \overleftarrow{h_{1}} \\
    \overrightarrow{h_{1:N}} &= \overrightarrow{Transformer}(s_1^r,s_2^r,\ldots s_N^r) \\
    \overleftarrow{h_{1:N}} &= \overleftarrow{Transformer}(s_N^r,s_{N-1}^r,\ldots s_1^r) 
\end{split}
\end{gather}
\subsection{Joint Layer and Loss Function}
We use a flexible way to fuse the output vectors $h^w$, $h^l$ and $h^d$ respectively obtained from above three subnetworks. Each subnetwork separately adjusts the weight of each neuron in its output vector through a fully connected layer, producing three intermediate vectors. Then we cascade them to obtain the final output vector $h^f$ to predict the classification result. Finally, we apply a fully connected layer with $sigmoid$ non-linear activation on $h^f$ to get a probability distribution $P_t$. The corresponding category with 
the maximum probability is taken as the predicted result $\hat y$ .




Our loss function is the cross entropy loss for the binary classification, which is defined as follow:
\begin{equation}
    L = -{\frac{1}{M^{'}}} {\sum_{i = 1}^{M^{'}}}(y_ilog(\hat{y_i})+ (1-y_i)log(1-\hat{y_i}))
\end{equation}
where $y_i$ and $\hat{y_i}$ are the ground truth and predicted probability of $i^{th}$ training sample respectively, and $M^{'}$ is the total number of training samples.
\begin{table}
	\centering
	\caption{\label{tab:01}Dataset Statistics. MSL is the mean sentence length, and MSN denotes the average number of sentences.}
	\small
	\resizebox{.95\columnwidth}{!}{\begin{tabular}{m{2.0cm}<{\centering}m{1.0cm}<{\centering}m{1.0cm}<{\centering}m{1.0cm}<{\centering}m{1.0cm}<{\centering}}
		\hline
		\rule{0pt}{0.3cm} Data & Positive & Negative& MSL & MSN \\
		\hline
		\rule{0pt}{0.3cm}Training set &20,888&15,360&135&38 \\
		\hline
		\rule{0pt}{0.3cm}Validation set &592&408&138&36 \\
		\hline
		\rule{0pt}{0.3cm}Test set &574&426& 134&39 \\
		\hline
	\end{tabular}}
\end{table}
\section{Experiment}
\subsection{Dataset}
For evaluation purposes, since there is no public benchmark dataset for our task yet, we construct a self-media online article quality classification dataset\footnote{https://drive.google.com/drive/folders/1VRNm5I-ZahfDlSH7gbfkhXcPJ69vhGDy} from Wechat, a well-known mobile self-media platform in China, where both media organizations and personal users can set up their official accounts for publishing news and articles\citep{LiaoXLHLL19}. Our dataset covers 44 categories of articles on this platform, including news, finance, technology, people's livelihood, etc. When reading articles, users can take ``share'', ``save'', ``like'' and ``tip'' actions. High-quality articles tend to be more popular and widely spread by users. The professional level of an author account also affects the quality of articles, which considers the authority and originality of the published articles on the account. Since the proportion of high-quality articles is only about 20\% of total articles per day, to narrow down the search scope, we first collect articles with high ``view'', ``like'', ``share'' and high author account level as high-quality candidates. Conversely, articles with low popularity and low author account level are selected as low-quality candidates. Note that the above is just the data collection process, not the results of annotation.

Our dataset is ultimately manually reviewed and annotated according to the criteria in Section 3.1 for discriminating high-quality and low-quality articles. We hired an annotating team to complete the dataset annotation. We divided the huge dataset into 10 subsets and each subset was labeled by two annotators. Each annotator was tested to label an additional 100 samples, and only qualified people were employed. The average Kappa coefficient of annotation consistency is 0.687. We only reserved samples with the consistent annotated label. Finally, we get 38,248 articles in our dataset, containing 22,054 high-quality articles as positive samples and 16,194 low-quality articles as negative samples. We respectively randomly extract 1000 samples from the dataset as the validation set and test set. Some statistics of the dataset are shown in Table \ref{tab:01}. 
\subsection{Baselines}
We compare our proposed model with following baselines. For brevity, $\overline{\mathcal{L}}$ denotes LSTM and $\overline{\mathcal{C}}$ represents CNN in model names.
\begin{itemize}
\item \textbf{Feature-based classiﬁers:} We adopt Logistic Regression (LR) and Random Forest (RF) as baselines. These classiﬁers take all the writing characteristics features as inputs.
\item \textbf{TextCNN:} A popular CNN-based text classifier \citep{Kim14}. 
\item \textbf{LSTM/ BiLSTM/ LSTM-att/ BiLSTM-att:} Unidirectional and bidirectional LSTM, and the variant models with the attention mechanism \citep{BahdanauCB14}. 
\item \textbf{Hierarchical Attention Networks:} The best performing network structure for the text quality assessment. We compare with many models with different sentence-level and document-level structures. \textbf{$\overline{\mathcal{C}}$-$\overline{\mathcal{L}}$-att} denotes the network whose sentence-level encoder is CNN and document-level encoder is LSTM. Other models' naming principles are the same. 
\item \textbf{Qe-HCRN:} A qualitatively enhanced convolutional recurrent neural network\citep{DasguptaNDS18} that incorporates different complex linguistic, cognitive and psychological features, joining two hierarchical convolutional recurrent networks to receive text and feature inputs respectively.
\item \textbf{Vani. BERT:} Replace the hi-BERT in our CoQAN with the vanilla non-hierarchical BERT.
\end{itemize}

\begin{table}
	\centering
		\caption{\label{tab:02} Overall performance of comparison models. Here $Acc.$ denotes accuracy and $Prec.$ denotes precision. $\overline{\mathcal{L}}$ denotes LSTM and $\overline{\mathcal{C}}$ represents CNN. $*p\le 0.01$ indicates that results are statistically signiﬁcant.
}
	\small
	\resizebox{.95\columnwidth}{!}{\begin{tabular}{m{1.7cm}<{\centering}m{0.8cm}<{\centering}m{0.8cm}<{\centering}m{0.8cm}<{\centering}m{0.8cm}<{\centering}m{0.8cm}<{\centering}}
		\hline
		\rule{0pt}{0.4cm} \textbf{Model}&\textbf{Acc.} &\textbf{Prec.}&\textbf{Recall}&\textbf{F1}&\textbf{AUC} \\
		\hline
		\rule{0pt}{0.3cm}LR &0.729&0.7299&0.7290&0.7221&0.7083 \\
		\hline
		\rule{0pt}{0.3cm}RF &0.837&0.8387&0.8370&0.8350&0.8247 \\
		\hline
		\rule{0pt}{0.3cm}Text$\overline{\mathcal{C}}$&0.705&0.7088&0.7050&0.6927&0.7376 \\
		\hline
		\rule{0pt}{0.3cm}$\overline{\mathcal{L}}$ &0.669&0.6653&0.6690&0.6611&0.7230 \\
		\hline
		\rule{0pt}{0.3cm}Bi$\overline{\mathcal{L}}$ &0.642&0.6634& 0.6420&0.5951&0.7038 \\
		\hline
		\rule{0pt}{0.3cm}$\overline{\mathcal{L}}$-att &0.649&0.6766& 0.6490&0.6020&0.6882 \\
		\hline
		\rule{0pt}{0.3cm}Bi$\overline{\mathcal{L}}$-att &0.680&0.6922& 0.6800&0.6560&0.7273 \\
		\hline
		\rule{0pt}{0.3cm}$\overline{\mathcal{L}}$-$\overline{\mathcal{L}}$-att &0.877&0.8776& 0.8770&0.8762&0.9417 \\
		\hline
		\rule{0pt}{0.3cm}Bi$\overline{\mathcal{L}}$-Bi$\overline{\mathcal{L}}$-att &0.878&0.8783& 0.8780&0.8773&0.9415 \\
		\hline
		\rule{0pt}{0.3cm}$\overline{\mathcal{C}}$-$\overline{\mathcal{C}}$-att &0.816&0.8186& 0.8160&0.8131&0.8764 \\
		\hline
		\rule{0pt}{0.3cm}$\overline{\mathcal{L}}$-$\overline{\mathcal{C}}$-att &0.845&0.8463& 0.8450&0.8434&0.9121 \\
		\hline
		\rule{0pt}{0.3cm}Bi$\overline{\mathcal{L}}$-$\overline{\mathcal{C}}$-att &0.851&0.8507& 0.8510&0.8508&0.9166 \\
		\hline
		\rule{0pt}{0.3cm}$\overline{\mathcal{C}}$-$\overline{\mathcal{L}}$-att &0.871&0.8714& 0.8710&0.8702&0.9358 \\
		\hline
		\rule{0pt}{0.3cm}$\overline{\mathcal{C}}$-Bi$\overline{\mathcal{L}}$-att &0.882&0.8821& 0.8820&0.8814&0.9433 \\
		\hline 
		\rule{0pt}{0.3cm}Qe-HCRN &0.904&0.9041& 0.9040&0.9041&0.9575 \\
		\hline 
		\rule{0pt}{0.3cm} Vani. BERT &0.907&0.9081& 0.9070&0.9072&0.9680 \\
		\hline
		\rule{0pt}{0.3cm}\textbf{CoQAN*} &\textbf{0.924}&\textbf{0.9241}& \textbf{0.9240}&\textbf{0.9240}&\textbf{0.9763} \\
		\hline
	\end{tabular}}
\end{table}
\subsection{Parameter Settings}
We initialize the weight matrices from the Gaussian stochastic distribution $\mathcal{N}(0,0.01)$ while all the bias vectors are set to $ \mathbf{0} $. In the layout organization subnetwork, the maximum sequence length of content blocks is 256 and the layout feature number for input vectors is 13. We adopt a single GRU layer with hidden size 128. We employ 4 kinds of kernels in CNN with sizes of 2,5,10,20 respectively. The number of each kind of kernels is set as 25. In the writing characteristics subnetwork, we apply 3 multi-head self-attention layers with 4 heads in each layer and hidden size 64 for a head, and the embedding size of features is 128. In the text semantics subnetwork, the maximum word number in a sentence is 128 and maximum sentence number is 32. The document-level bidirectional Transformer's parameters are the same with the BERT blocks at the sentence level, with 12 attention heads, attention hidden size 768 and intermediate hidden size 3072. The vocabulary is the one that comes with BERT, whose length is 21,128. We start with the pre-trained BERT at the sentence level and fine-tune its parameters with our dataset. The fully connected layers for the subnetwork fusion have hidden size 128 for the text semantics subnetwork and hidden size 64 for the other two subnetworks. Besides, we adopt two Adam optimizers with different learning rates to update parameters in the joint training, learning rate $2e^{-5}$ and dropout 0.1 for BERT fine tuning, while learning rate 0.001 and dropout 0.2 for other parts, with a batch size of 8. We use the early stop training strategy to prevent overfitting. All baseline models are implemented with the same settings for the same parts.

\begin{table}
	\centering
	\caption{\label{tab:03} Comparison of network structures for the layout organization subnetwork. $\overline{\mathcal{G}}$ for GRU and $\overline{\mathcal{C}}$ for CNN.}
	\small
	\resizebox{.95\columnwidth}{!}{\begin{tabular}{m{1.1cm}<{\centering}m{0.8cm}<{\centering}m{0.8cm}<{\centering}m{0.8cm}<{\centering}m{0.8cm}<{\centering}m{0.8cm}<{\centering}}
		\hline
		\rule{0pt}{0.4cm} \textbf{Model}&\textbf{Acc.} &\textbf{Prec.}&\textbf{Recall}&\textbf{F1}&\textbf{AUC} \\
		\hline
		\rule{0pt}{0.3cm}$\overline{\mathcal{G}}$&0.775&0.7743&0.7750&0.7727&0.8369 \\
		\hline
		\rule{0pt}{0.3cm}Bi$\overline{\mathcal{G}}$ &0.774&0.7731& 0.7740&0.7721&0.8463 \\
		\hline
		\rule{0pt}{0.3cm}$\overline{\mathcal{C}}$&0.778&0.7798& 0.7780&0.7739&0.8435 \\
		\hline
		\rule{0pt}{0.3cm}Bi$\overline{\mathcal{G}}$-$\overline{\mathcal{C}}$ &0.781&0.7836& 0.7810&0.7766&0.8368 \\
		\hline
		\rule{0pt}{0.3cm}\textbf{$\overline{\mathcal{G}}$-$\overline{\mathcal{C}}$} &\textbf{0.785}&\textbf{0.7847}& \textbf{0.7850}&\textbf{0.7827}&\textbf{0.8485} \\
		\hline
	\end{tabular}}
\end{table}
\subsection{Result Analysis}
\subsubsection{Comparison with Baselines}
Table \ref{tab:02} shows that compared with many baseline models, our proposed method CoQAN achieves the best results for all metrics on the test set. Most previous methods for assessing article quality are based solely on text modeling. The single level BiLSTM and LSTM models get the worst results, since the text input of an article is too long, it is so difficult to capture such a long distance dependency. The performance of TextCNN is a little better, probably it captures some local keywords or key fragments that contribute to the quality of articles. Feature-based classiﬁers are better than basic textual models in our task, which also reveals that single level recurrent networks and convolutional networks are difficult to capture the complicated features and dependencies in long article text. Hierarchical attention networks get much more better performance, indicating that the hierarchical structure is more suitable for handling inter-word and inter-sentence relations in document-level text inputs. From the results, the performance of bidirectional networks are usually better than the unidirectional networks, regardless of the sentence level or the document level. When applying CNN at the sentence level and BiLSTM at the document level, it gets the best performance among these hierarchical models. This is likely because that local n-gram information is more relevant to the scoring of sentence structures, while global information is more relevant for scoring document-level coherence \citep{DongZY17}. The Qe-HCRN augments specific qualitative features in the hierarchical attention neural network, thus it performs better than feature-based baseline models and hierarchical attention networks. Since the Qe-HCRN model uses all the same features as ours, the results prove that our model's superior performance benefits from the network architecture we designed, not the addition of more features. Our model CoQAN significantly exceeds all baselines, indicating the effectiveness of our joint learning of multi-space subnetworks, which can take advantage of more valuable information. Besides, our model is better than Vani. BERT, which proves the validity of the hi-BERT we designed.

We also conduct comparative experiments to discuss the network structure of the layout organization subnetwork, the results are shown in Table \ref{tab:03}. The performance of cascade networks are better than that of single networks, which proves the superiority of learning global sequence dependency and local patterns simultaneously. In addition, since there are many features calculated from the top of the page in the layout feature vectors, it is expected that GRU performs better than BiGRU.

\begin{table}
	\centering
	\caption{\label{tab:04} Results of ablation experiments. LO, WC, and TS represent the layout organization, writing characteristics, and text semantics subnetworks respectively.}
	\small
	\resizebox{.95\columnwidth}{!}{\begin{tabular}{m{1.1cm}<{\centering}m{0.8cm}<{\centering}m{0.8cm}<{\centering}m{0.8cm}<{\centering}m{0.8cm}<{\centering}m{0.8cm}<{\centering}}
		\hline
		\rule{0pt}{0.4cm} \textbf{Model}&\textbf{Acc.} &\textbf{Prec.}&\textbf{Recall}&\textbf{F1}&\textbf{AUC} \\
		\hline
		\rule{0pt}{0.3cm}LO &0.785&0.7847& 0.7850&0.7827&0.8485 \\
		\hline
		\rule{0pt}{0.3cm}WC &0.850&0.8499&0.8500&0.8492&0.9178 \\
		\hline
		\rule{0pt}{0.3cm}TS &0.895&0.9004& 0.8950&0.8934&0.9584 \\
		\hline
		\rule{0pt}{0.3cm}LO-WC &0.873&0.8740& 0.8730&0.8720&0.9255 \\
		\hline
		\rule{0pt}{0.3cm}LO-TS &0.904&0.9049& 0.9040&0.9042&0.9667 \\
		\hline
		\rule{0pt}{0.3cm}WC-TS &0.915&0.9174& 0.9150&0.9142&0.9701 \\
		\hline
		\rule{0pt}{0.3cm}\textbf{CoQAN}  &\textbf{0.924}&\textbf{0.9241}& \textbf{0.9240}&\textbf{0.9240}&\textbf{0.9763}\\
		\hline
	\end{tabular}}
\end{table}

\subsubsection{Ablation Analysis}
We conduct the ablation study to examine the effect of each subnetwork, the results are shown in Table \ref{tab:04}. The results show that when only reserving one subnetwork, the performance of the text semantics subnetwork is best, and the performance of our network CoQAN decreases most when the text semantics subnetwork is eliminated. This proves that modeling complex writing knowledge is necessary and effective. Compared with the results in Table \ref{tab:02}, we can see that the performance of our only text semantics subnetwork exceeds all the baselines which only model the text input, and the performance of our writing characteristics subnetwork surpasses all the feature-based baseline models. These comparison results indicate the superiority of the designed architectures in our subnetworks. In Table \ref{tab:04}, the results prove that 
the layout quality modeling can help improve the performance of the article quality assessment. In addition, the performance of any two joint subnetworks is better than a single subnetwork, and the whole network CoQAN joining three subnetworks is the best, which shows the effectiveness of our joint learning method, and each subnetwork contributes to the performance improvement of the self-media online article quality classification.

\begin{figure}
\centerline{\includegraphics[width=\columnwidth,height=5.3cm]{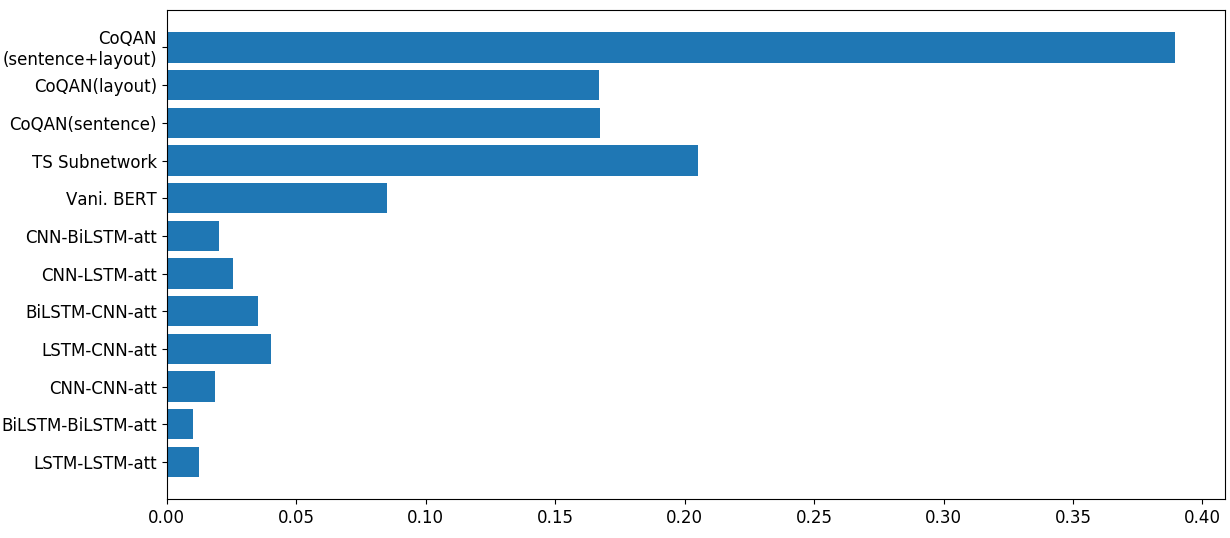}}
\caption{Results of disturbance experiments.}
\label{figure4}
\end{figure}

\begin{figure*} 
	\centering
	\subfigure[Self-attention layer 1 
]{
		\includegraphics[width=0.65\columnwidth,height=5cm]{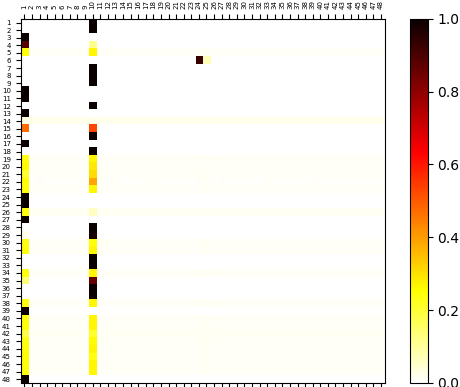}
		\label{layer1}
	}
		\subfigure[Self-attention layer 2 
]{
		\includegraphics[width=0.65\columnwidth,height=5cm]{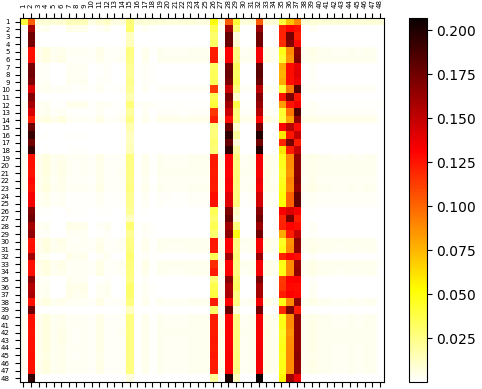}
		\label{layer2}
	}
		\subfigure[Self-attention layer 3  
]{
		\includegraphics[width=0.65\columnwidth,height=5cm]{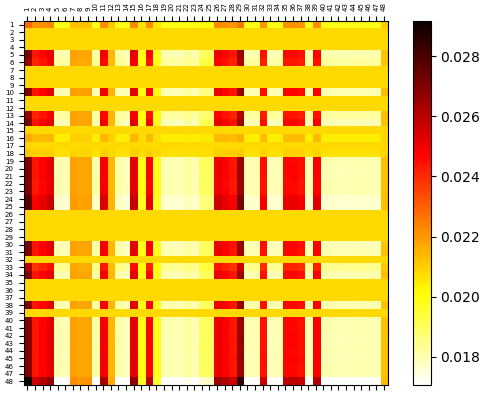}
		\label{layer3}
	}
	\caption{Visualization results of different self-attention layers in our writing characteristics subnetwork.}
	\label{figure5}
\end{figure*}
\subsubsection{Disturbance Analysis}
We also design disturbance experiments to verify whether the models can learn the writing logic and layout patterns we care about. Here, we randomly shuffle the order of sentences or layout content blocks of all correctly predicted positive samples. Then we regard the shuffled samples as negative samples, feed them into comparison models and count the proportion of the correct prediction as negative samples, that is, the proportion of successful disturbances. In order to eliminate the effect of randomness, each of our shuffle experiments is repeated 5 times and the average values are taken as the final results. 

Figure \ref{figure4} shows the successful disturbance rates of sentence shuffle among comparison models. From the results, our model CoQAN is obviously more sensitive to the disturbance of sentence disordering than all baselines, which shows that CoQAN can easier capture the cohesive relationships between sentences and learn the consistency of content and writing logic. Note that the successful disturbance rate of sentence shuffle in the text semantics (TS) subnetwork is higher than that in our whole joint network CoQAN, this is because we conduct all disturbance experiments on positive samples, and positive samples have both good text quality and layout quality. Thus, when only disordering sentences, good layouts prevent the CoQAN to judge them as negative samples to a certain extent. Figure \ref{figure4} also shows the successful disturbance rates of different shuffle dimensions of CoQAN. When we shuffle the layout order and sentence order at the same time, the proportion of successful disturbances is much higher than the disturbance under a single dimension. All of these results prove the effective integration of the layout learning and text learning in our proposed model, and our model successfully learns the knowledge of both text writing and page layouts.

\subsubsection{Feature Study}
In order to study the behavior of different features and the process of feature interactions, we visualize the attention weights of different self-attention layers in our writing characteristics subnetwork in Figure \ref{figure5}, and give some qualitative analysis. In our dataset, the text in articles is mainly in Chinese. The numbers on the coordinate axis in Figure \ref{figure5} are the index values of features. The mapping between the feature names and their indexes is in Appendix A. Form Figure \ref{figure5}, we find that the bottom attention layers learn dominant features, while the top layers learn more complex feature spaces and intricate interactions. This proves that our writing characteristics subnetwork can sufficiently learn the interactions between different features and obtain meaningful high-order combinatorial features. Figure \ref{layer1} shows that the most dominant features are the text length (10) and the maximum number of characters in pictures (1). The numbers in the brackets here are the index values on the coordinate axis in Figure \ref{figure5}. This is because that high-quality articles usually have richer content and regular pictures, while low-quality articles usually have incomplete content, and pictures with much text are very likely to be irrelevant advertising pictures, also affecting readability. Other important features include the maximum proportion of the textual area in pictures (2), the unique word number (15), keyword number in the title (28), picture number (32), paragraph number (35) and pos tags (36,37), suggesting that the quality of pictures, titles, writings and body organizations are all important for self-media online articles. 

\section{Conclusion}
In this work, we propose a novel method for the self-media online article quality classification. We design a joint network CoQAN to decouple the modeling of the layout organization, writing characteristics and text semantics, finally merge them into a unified model. We innovatively propose to explicitly learn the presentational quality of online articles, together with the text quality to predict the final quality of online articles. The proposed framework can integrate different features of the online article quality assessment, and the specially designed high-order interactive feature learning and the mobile browsing habits modeling can solve the problems in this field well. Evaluation results based on the real-world online article dataset demonstrate the effectiveness and superiority of our proposed CoQAN. Since most self-media articles convey the main content and core ideas mainly through text, and considering the complexity of this task and the efficiency in practical applications, our proposed network focuses more on understanding the semantics of the text. More generalized assessments may need to introduce the semantic judgment of pictures and we leave this into our future work.

\appendix

\section{Writing Features}

Here we list the feature names and their indexes used in Figure \ref{figure5}. The numbers in the brackets are the index values. 

Maximum character number in pictures (1), Maximum proportion of the textual area in pictures (2), Number of valid pictures (Remove template pictures, emoticons and QR code pictures) (3), Ratio of the pictures containing text to the total pictures (4), Number of template pictures (Usually as the signal of section divisions) (5), Total character number (including the text in pictures) (6), Number of pictures containing text (7), Total word number (8), Character number after removing stop words (9), Total character number in text (10), Unique character number (11), Word number after removing stop words (12), Unique word number (13), Ratio of unique character number to total character number (14), Ratio of unique word number to total word number (15), Punctuation number (16), Noun number (17), Verb number (18), Adjective number (19), Ratio of punctuation (20), Ratio of nouns (21), Ratio of verbs (22), Ratio of adjectives (23), Title length (24), Title length after removing stop words (25), Ratio of the title length after removing stop words to the total title length (26), Word number in the title (27), Keyword number in the title (28), Ratio of keywords to total words in the title (29), Ratio of picture number to word number (30), GIF number (31), Image number (exclude GIF) (32), Total picture number (33), Video number (34), Paragraph number (35), Number of conjunctions (36), Number of pronouns (37), Number of adverbs (38), Number of numerals (39), Number of auxiliary words (40), Number of idioms (41), Ratio of paragraph number to picture number (42), Ratio of conjunctions (43), Ratio of adverbs (44), Ratio of numerals (45), Ratio of auxiliary words (46), Ratio of idioms (47), Article category (48). 








\begin{acks}
This research was partially supported by the Key Program of National Natural Science Foundation of China under Grant No. U1903213, the Guangdong Basic and Applied Basic Research Foundation (No. 2019A1515011387), the Dedicated Fund for Promoting High-Quality Economic Development in Guangdong Province (Marine Economic Development Project: GDOE[2019]A45), and the Shenzhen Science and Technology Project under Grant (ZDYBH201900000002, JCYJ20180508152042002).
\end{acks}

\bibliographystyle{ACM-Reference-Format}
\balance
\bibliography{sample-base}

\end{document}